\documentclass{bmvc2k}

\title{Log NeRF: Comparing Spaces for Learning Radiance Fields}

\addauthor{Sihe Chen}{chen.sihe1@northeastern.edu}{}
\addauthor{Luv Verma}{verma.lu@northeastern.edu}{}
\addauthor{Bruce A. Maxwell}{b.maxwell@northeastern.edu}{}

\addinstitution{
 Northeastern University \\
 Boston, MA, USA
}

\runninghead{Chen, Verma, Maxwell}{Log NeRF}

\usepackage{graphicx}
\usepackage{makecell}
\usepackage{multirow}

\begin{document}

\maketitle

\begin{abstract}
Neural Radiance Fields (NeRF) have achieved remarkable results in novel view synthesis, typically using sRGB images for supervision. However, little attention has been paid to the color space in which the network is learning the radiance field representation. Inspired by the Bi-Illuminant Dichromatic Reflection (BIDR) model, which suggests that a logarithmic transformation simplifies the separation of illumination and reflectance, we hypothesize that log-RGB space enables NeRF to learn a more compact and effective representation of scene appearance.

To test this, we captured approximately 30 videos using a GoPro camera, ensuring linear data recovery through inverse encoding. We trained NeRF models under various color space interpretations—linear, sRGB, GPLog, and log RGB—by converting each network output to a common color space before rendering and loss computation, enforcing representation learning in different color spaces. Quantitative and qualitative evaluations demonstrate that using a log RGB color space consistently improves rendering quality, exhibits greater robustness across scenes, and performs particularly well in low-light conditions while using the same bit-depth input images. Further analysis across different network sizes and NeRF variants confirms the generalization and stability of the log-space advantage.
\end{abstract}

\section{Introduction}

Neural Radiance Fields (NeRF) are a popular technique for novel view synthesis and 3D scene representation~\cite{mildenhall-nerf-2020}. NeRF methods use a multi-layer perceptron (MLP) to learn a continuous representation of a scene's volumetric structure and appearance from a set of training images. For a single query, the input to the NeRF network is a 3D coordinate ($x$, $y$, $z$) and a 2D viewing direction ($\theta$, $\phi$), typically combined with a positional encoding. The MLP then outputs a single color value (R, G, B) and a scalar volume density $\sigma$ for that specific point and direction. To render a novel high-resolution view, a volumetric renderer will sample multiple points along each camera ray, querying the network at each point to gather a set of colors and densities. It then integrates the values along the ray to compute a final color for each pixel.

There are many variations on NeRF models that have explored a wide variety of issues that arise when trying to model the light field of a scene. With a few exceptions that have used RAW image data (e.g. \cite{mildenhall2022nerf}), NeRF work to date has focused on either synthetic data or data captured with consumer devices and processed for human viewing. The implied assumption in most prior work is that the network representing the light field is learning it in the input color space, also assumed to be the output color space, which is generally sRGB.

%The sRGB color space is the result of a non-linear compression of linear RGB space in order to brighten shadows and make images look more normal to human viewers. The non-linear sRGB transformation causes the rules that govern the interaction of light and matter to become color dependent and more complex than in linear space or spaces created by transformations that respect the physics of the scene \cite{maxwell2008bi}.

One reason that most NeRF work has used videos processed for human viewing is that obtaining linear RGB videos has been challenging without high-end video cameras. While capturing single images in RAW mode is possible with most consumer devices, capturing RAW video has not been as simple (though it is becoming easier \cite{motioncam-app}). Recently, however, GoPro released consumer cameras that can capture video in their GPLog encoding format, which is a 10-bit transformation of the linear RGB data. The transformation is invertible into linear RGB without significant loss, which enables transformations into other spaces like log RGB and sRGB. This makes it possible to easily capture linearizable videos appropriate for NeRF processing.  While more time-consuming than capturing a scene with video, it is also possible to linearize a large set of photos of a scene taken with any camera that can capture RAW data. However, pose estimation can be more difficult compared to video sequences.

In this work, we explore whether the representation space in which the network learns the radiance field impacts the training, compactness, robustness, dynamic range, and quality of the NeRF process. For each of these characteristics, we execute a series of experiments and compare the GPLog encoding, sRGB, linear RGB, and log RGB. In order to achieve fair comparisons, the training data in all cases is sRGB derived from the GPLog encoded video, and losses are computed in the sRGB color space.

To make our work replicable, we use the BiLaRF NeRF model as the basis for all of our experiments~\cite{wang2024bilateral}. We control the radiance field color space implicitly using two small modifications. First, we use a hard-coded transformation from the implicit color space to linear space between the radiance field MLP and the differentiable renderer. We then execute an sRGB transform on the linear renderer output in order to compute losses in sRGB.

We explore compactness by evaluating the relationship between the learned space, network size, network training time, and Peak Signal-to Noise Ratio (PSNR). We explore robustness by executing multiple training runs using each color space and computing statistics on performance. We compare dynamic range by evaluating each space on a video with controlled variation in intensity. Finally, we compare quality through PSNR for the output images and a qualitative comparison of the depth maps learned by the network on a range of real videos taken indoors and outdoors with both simple and complex lighting situations.

In all of these metrics we found that log RGB outperformed the other color spaces, in many cases by a significant margin using the same base architecture and training procedure. In order to determine the extent to which log RGB is special, we also tested similar, but not exactly log functions. Only true log RGB demonstrated improvement over the other spaces.

Overall, our contributions are as follows: (1) we explore the impact of the internal representation space for radiance fields, (2) we demonstrate how to control the implicit representation space of a NeRF network, (3) we show that log RGB outperforms other color spaces on multiple metrics, (4) we show that log RGB is special; similar transforms do not show the same benefits, and (5) we contribute a new set of NeRF videos in GPLog encoding format along with a process for linearizing them.

The idea of learning radiance fields in a different color space seems simple.  But the impacts of this small change are profound across multiple metrics and methods of evaluation.

\section{Related and Prior Work}

\textbf{Novel View Synthesis (NeRFs)} Novel view synthesis aims to construct scene representations capable of rendering new viewpoints from a set of images and camera poses. Early methods relied on dense sampling and direct pixel interpolation~\cite{gortler1996lumigraph}~\cite{levoy1996light}. Recent deep learning approaches have shifted towards volumetric scene representations~\cite{flynn2016deepstereo,lombardi2019neural,zhou2018stereo}. Neural Radiance Fields (NeRF)~\cite{mildenhall-nerf-2020} pioneered the use of neural volumetric scene representations optimized through gradient descent on a rendering loss. Subsequent extensions enhanced NeRF's capabilities, addressing varying lighting conditions~\cite{martin2021nerf}, incorporating depth~\cite{deng2021depth,jeong2021self,wei2021nerfingmvs}, utilizing time-of-flight data~\cite{attal2021torf}, and integrating semantic segmentation~\cite{zhi2021place}.

A recent innovation in NeRF is the use of a 3D bilateral grid to model per-frame camera processing such as gain or color balance changes representable using a 3x4 affine transformation of the color space~\cite{wang2024bilateral}. 
%These transformations are not intended to model the average ISP processing--such as conversion to sRGB--but per-frame modifications, enabling high quality renderings even when the intensity or color balance of the frame sequence is inconsistent.  The NeRF model itself must learn the average ISP processing. 
The BiLaRF implementation is publicly available and we use it as the basis for our analysis of color spaces. However, we actually turn off the 3-D bilaterial grid used during training to avoid clipping, simplifying the model.

RawNeRF~\cite{mildenhall2022nerf} has demonstrated the potential of incorporating raw HDR data into NeRF frameworks. By training on raw linear data instead of tone-mapped sRGB images, RawNeRF can handle high dynamic range scenes and produce HDR novel views. This approach has shown remarkable robustness to high noise levels, effectively functioning as a multi-image denoiser for wide-baseline static scenes. We show that using log RGB doesn't require higher bit depths to achieve high quality results on darker sequences.

%The impressive results of the RawNeRF approach demonstrate the benefits of exploring alternative input spaces. sRGB is the most common color space used by consumer cameras, and it is optimized for human viewing of images on a computer. The linear data available from RAW images is a valuable alternative that offers higher dynamic range and preserves the physics of light-material interactions. However, it is not the only alternative color space. Furthermore, no one has directly compared the impact of using different color spaces with the same bit-depth of information.

Logarithmic RGB (log RGB) representations have been explored in various computer vision applications due to their illumination-invariant properties. In particular, differences in log space are ratios in linear space and are not sensitive to overall intensity. Log RGB has been utilized for illumination-invariant feature extraction~\cite{wang-logRGBHOG-2007}, material prior computation~\cite{put-materialPriors-BMVC2016}, and the creation of illumination-invariant color spaces~\cite{MarchantJOSA00}~\cite{FinlaysonECCV04}~\cite{maxwell2008bi}. These techniques have proven valuable in tasks such as skin lesion detection~\cite{polania-logRGBICA-2020}~\cite{madooei-skinLesion-2012}, color constancy~\cite{shi-colorConstancy-ECCV2016}, intrinsic image decomposition~\cite{Liu2018ConsistencyawareSO}, and shadow removal~\cite{maxwell-wad-2019}. The log transformation aligns with the human eye's logarithmic response to light \cite{Rose1948TheSP}, suggesting potential benefits for computational models.

Recent advancements have explored the use of log RGB in deep neural networks for classical vision tasks. Equivariant networks have been proposed to achieve invariance to input data offsets~\cite{cotogni-equivariantNetworks-2022}, while other studies have demonstrated the effectiveness of log RGB in providing robustness to intensity and color variations without modifying network architectures~\cite{Funt-LaplacianOfLog2022}~\cite{maxwell-2023-bmvc}~\cite{maxwell-logLenses-CVPR2024}. These findings suggest that log RGB representations can enhance the performance and generalization capabilities of deep learning models. However, to date, this has not been evaluated for learning radiance fields.

\section{Methods and Experimental Design}

\subsection{Data Capture and Pre-processing}

The videos were captured using a GoPro Hero 13 Black. Each video consists of a 360-degree rotation around a specific object, recorded under varying lighting conditions. In total, approximately 30 videos were collected for training and testing our method. Each video has a duration of 20–30 seconds.
To reverse-engineer the GoPro-log color representation, we referred to the official documentation on GPLog 10-bit encoding and validated our interpretation using a color checker. 

We used COLMAP to extract camera poses from video frames. However, for some very dark videos, COLMAP reported a low percentage of successfully registered frames. To address this, we converted the videos from GPLog to TrueLog color space, effectively brightening the images. Among the 13 videos with initially low pose recovery rates, 3 showed no improvement due to scenes being nearly completely dark, another 3 improved but still failed to reach the 100\% registration required by the network, and the remaining 7 videos successfully reached full registration after applying COLMAP in the log color space.

\subsection{Color Space Analysis}

Much of the light from a scene captured by a camera is reflected light that is a result of the interaction of illumination and materials. The Dichromatic Reflection Model, suitable for many natural and human-made materials, divides this into two types of reflection: body reflection and surface reflection~\cite{Shafer1985}. The Bi-Illuminant Dichromatic Reflection Model (BIDR) extends the Dichromatic model and separates the ambient light $A$ from the direct illuminant $D$, allowing them to be different colors, as occurs in most natural scenes (e.g. blue skylight and yellow sunlight)~\cite{maxwell2008bi}.

The BIDR model represents the body reflection as $I = R_{B}(A + \gamma D)$, where $I$ is the captured color, $R_{B}$ is the body reflectance, $A$ is the ambient illumination, $D$ is the direct illuminant, and $\gamma \in [0, 1]$ is the fraction of the direct illuminant incident on the surface point. In linear RGB space, the line or cylinder representing each material under varying illumination starts at a unique point $R_{B}A$, points in a unique direction defined by by $R_{B}D$, and is of a unique length, also defined by $R_{B}D$. Therefore, each material requires two colors to define it: $R_{B}A$ and $R_{B}D$, and these colors conflate material and illumination.

In log RGB space, however, the reflectance and illumination terms separate.
\begin{equation}
    \log I = \log R_{B}(A + \gamma D) = \log R_{B} + \log (A + \gamma D)
\end{equation}
The cylinder representing each material's body reflection under varying illumination still starts in a unique location. However, the illumination term remains the same across materials lit by the same ambient/direct illumination pair. All of the body cylinders in that set point in the same direction and are the same length. From a representational point of view, in log RGB each material is defined by a single point $R_{B}$, all materials in shadow share the offset $\log(A)$ and the body cylinders of all materials move from shadow to lit along the vector defined by $\log(A + \gamma D) - \log(A)$. Therefore, it is possible to represent all body reflection with one material color per pixel, one ambient color for the scene, and one illumination ratio vector for the scene, almost half the information required in linear space.

In sRGB space, which is approximated by a gamma-correction with a factor of 2.2 applied to linear RGB, the body reflection across varying illumination forms a curve, and each material requires learning a different curve of different length. As defining a curve is more complex than defining a line, we hypothesize that sRGB space requires more information to represent a single material under varying illumination than linear RGB.

The GoPro-log encoding is an intermediate space that is closer to linear than to sRGB or log RGB. As such, it likely requires at least as much information as linear RGB space in order to represent the appearance of a material under varying illumination.

\subsection{Network Modifications}

\begin{figure}[tbp]
    \centering
    \includegraphics[width=0.9\linewidth]{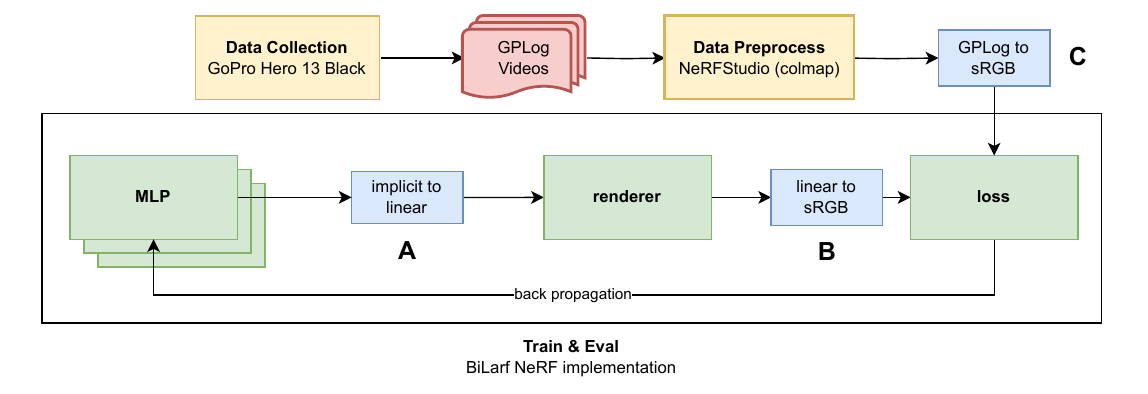}
    \caption{Modified BiLaRF NeRF architecture}
    \label{fig:network-diagram}
\end{figure}

We use the BiLaRF code base from \cite{wang2024bilateral} as the base architecture for all experiments. They all train using a batch size of 16,384, a Charbonnier loss function with significant Total Variation regularization, and a learning rate scheduled to decay from 0.01 to 0.001 over 25,000 epochs. As shown in Figure \ref{fig:network-diagram}, we added three components: (A) the representation space transform, (B) the sRGB transform, and (C) an input data transform.
The representation space transform converts from the implicit representation space learned by the NeRF network into linear RGB space, which is the most appropriate space for the differentiable renderer. The sRGB transform converts the linear RGB signal into sRGB. The input data transform modifies the input data format (e.g. GPLog or Linear) into sRGB using the standard sRGB transformation. 

We adopt four different color spaces as implicit representations within the network: GoPro-log (GPLog), linear RGB, sRGB, and true log RGB (TrueLog). All conversions between color spaces are performed in two steps: the source space is first transformed into linear RGB, followed by a mapping from linear RGB to the target space. Below, we define the forward transformations from linear RGB to each of the three non-linear spaces. The inverse transformations are simply the analytical inverses of these mappings.

GPLog: We approximate the logarithmic encoding used by GoPro cameras with (\ref{eq:gplog})

\begin{equation}
    {GPLog}(x) = \ln\left(x \cdot (e - 1) + 1\right)
    \label{eq:gplog}
\end{equation}

sRGB: We approximate the sRGB transformation using a gamma correction of 2.22.

\begin{equation}
{sRGB}(x) = x^{\frac{1}{2.22}}
\end{equation}

TrueLog: The TrueLog transformation is defined in (\ref{eq:truelog}).

\begin{equation}
f(x) = \ln\left( \frac{ e^{\ln(\max(255x, 1)) - 1} \cdot 255 }{ e - 1 } \right) %\approx \frac{ln(x)}{ln(255)} \text{ for } x \in [1, 255]
\label{eq:truelog}
\end{equation}

The TrueLog transform converts a linear signal in the range [0, 1] to a log signal in the range [0, 1]. To constrain the output of this function to the range $[0, 1]$ and avoid mathematical instability, we apply the further normalization step:
\begin{equation}
{TrueLog}(x) = \frac{f(x) - f(x_{\min})}{f(x_{\max}) - f(x_{\min})}
\end{equation}
where $x_{\min} = \frac{1}{255}$ and $x_{\max} = 1$ correspond to the lower and upper bounds of the input domain.  

% Figure \ref{fig:conversion} shows the conversion from linear to the four color spaces.

% \begin{figure}[t]
% \centering
% \includegraphics[height=1.25in]{images/conversion.pdf} \\
% \caption{Visualizations of color space transformations}
% \label{fig:conversion}
% \end{figure}

For all of our experiments we turned off the {\tt bilarf\_grid} flag and set the padding to zero to ensure the intermediate network results were in the range [0, 1] to facilitate the representation space transformations. The bilateral grid can cause clipping to occur, losing information and corrupting the implicit log representation of the network.

\section{Experiments and Results}

\textbf{Overall Quality:} We assess the quality of our network using PSNR, color renderings, and depth map visualizations. To balance reconstruction quality and training efficiency, we empirically found that 5000 training iterations provides consistently good results. To ensure robustness, we ran each experiment at least five times and report the best PSNR.

\begin{figure}[t]
\centering
\fontsize{8pt}{9pt}\selectfont
\begin{tabular}{|c|c|c|c|c|c|}
\hline
\multicolumn{1}{|c|}{\textbf{Color Space}} &
\multicolumn{1}{c|}{\textbf{GX010032}} &
\multicolumn{1}{c|}{\textbf{GX010103}} &
\multicolumn{1}{c|}{\textbf{GX010109}}\\ \hline
GPLog &
\includegraphics[width=0.23\linewidth]{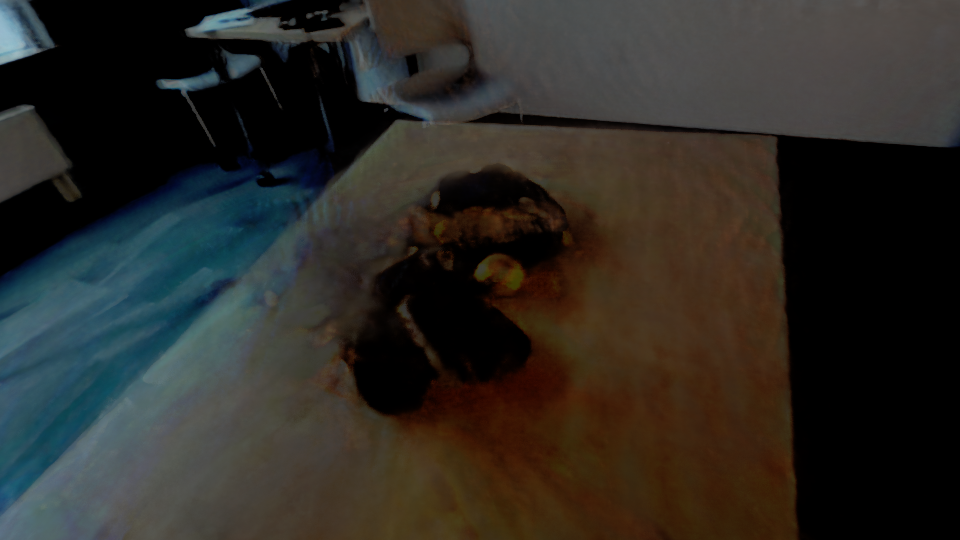} &
\includegraphics[width=0.23\linewidth]{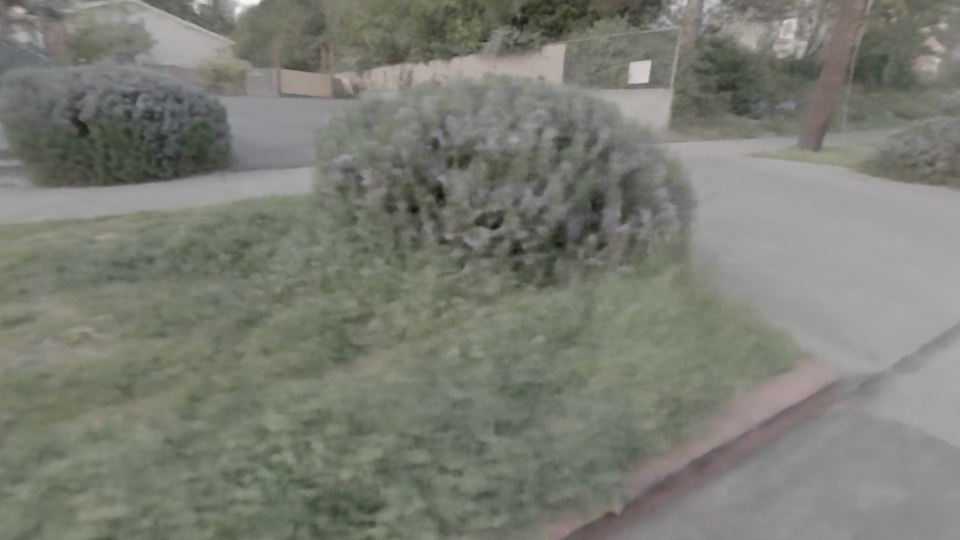} &
\includegraphics[width=0.23\linewidth]{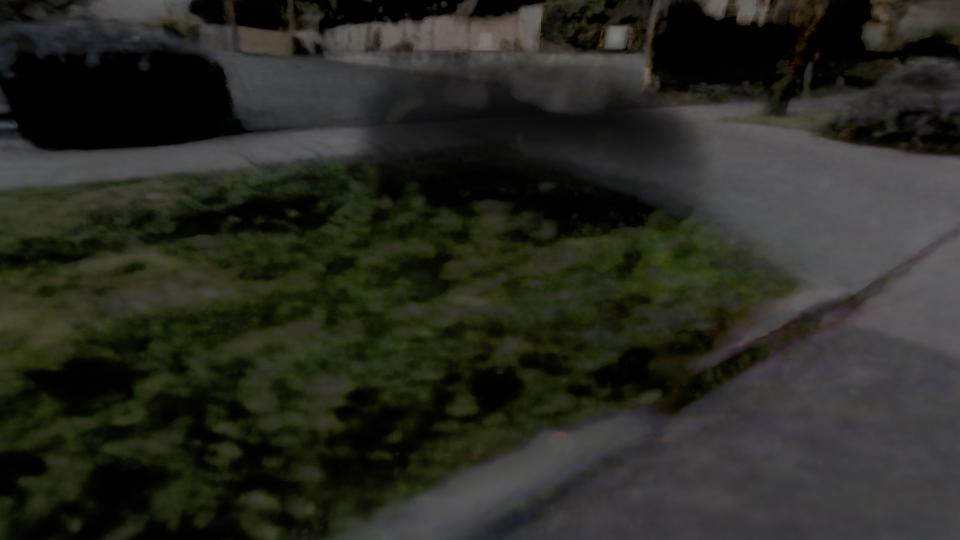} \\ \hline
linear &
\includegraphics[width=0.23\linewidth]{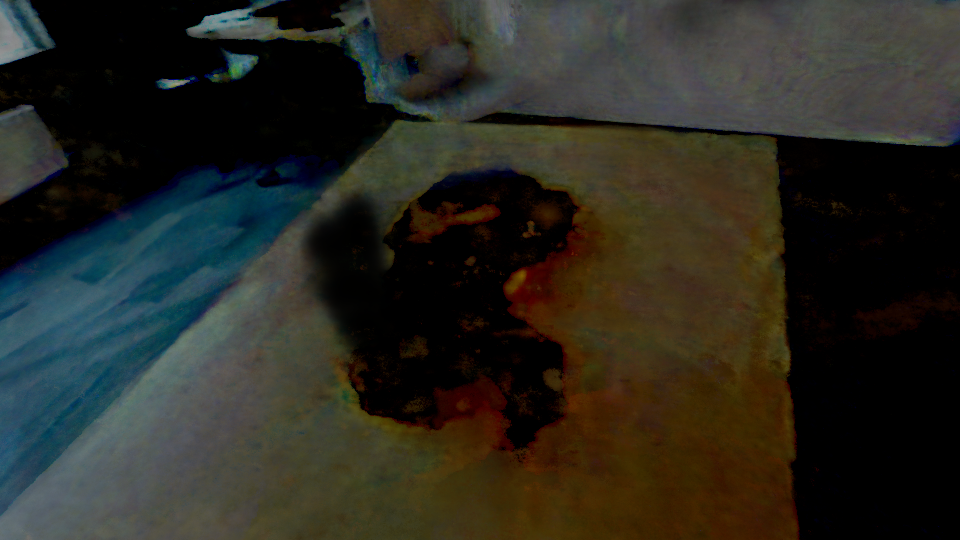} &
\includegraphics[width=0.23\linewidth]{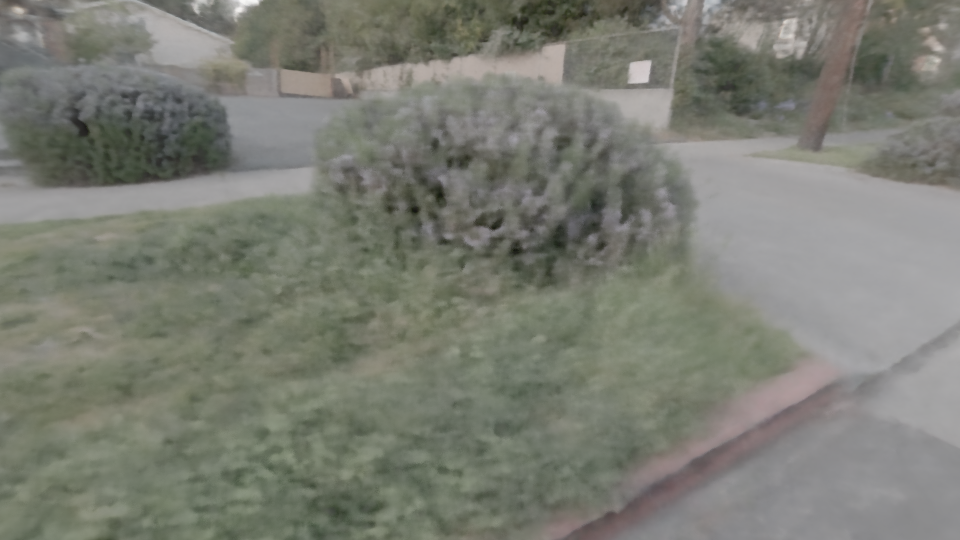} &
\includegraphics[width=0.23\linewidth]{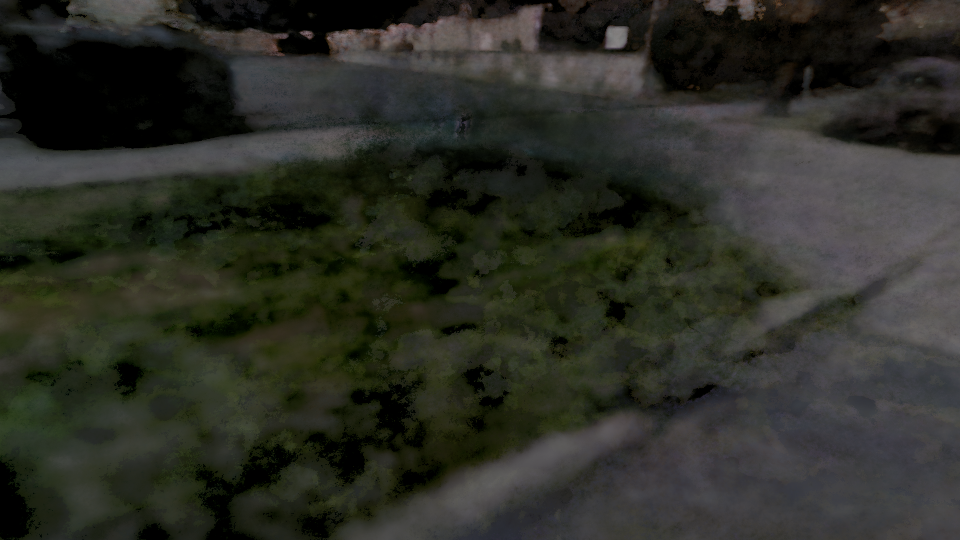} \\ \hline
sRGB &
\includegraphics[width=0.23\linewidth]{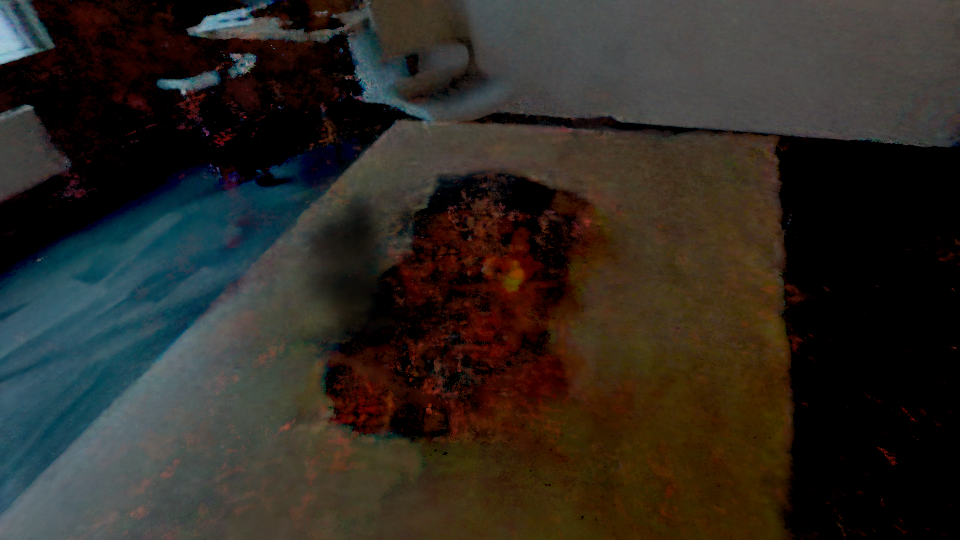} &
\includegraphics[width=0.23\linewidth]{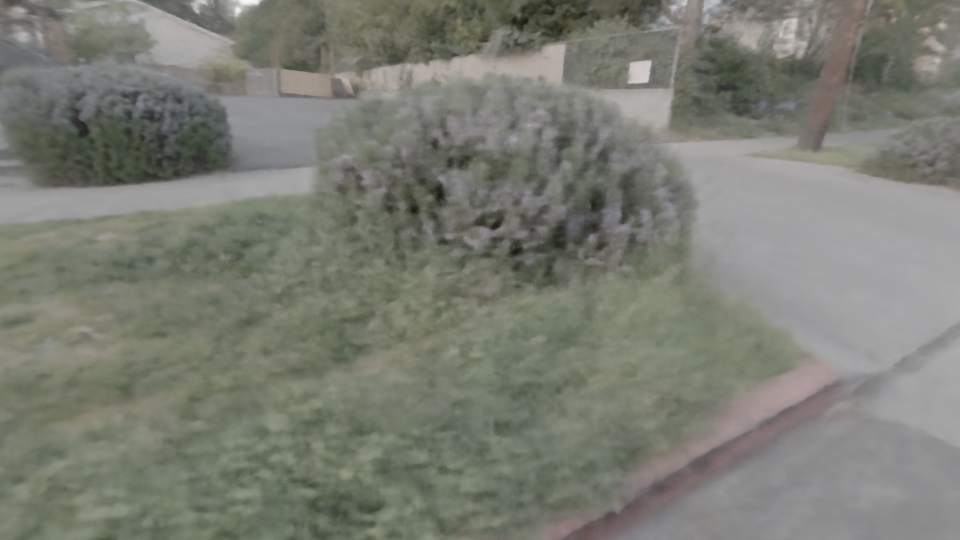} &
\includegraphics[width=0.23\linewidth]{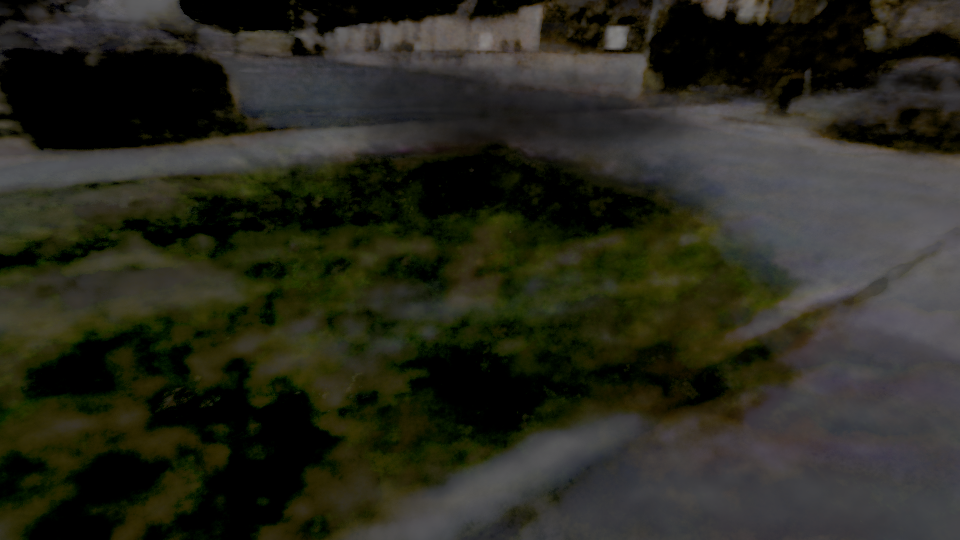} \\ \hline
TrueLog &
\includegraphics[width=0.23\linewidth]{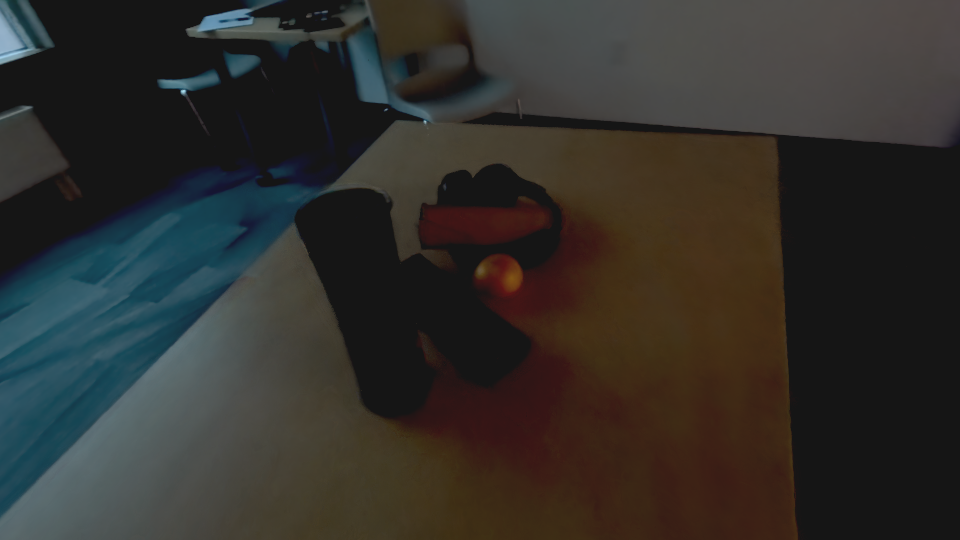} &
\includegraphics[width=0.23\linewidth]{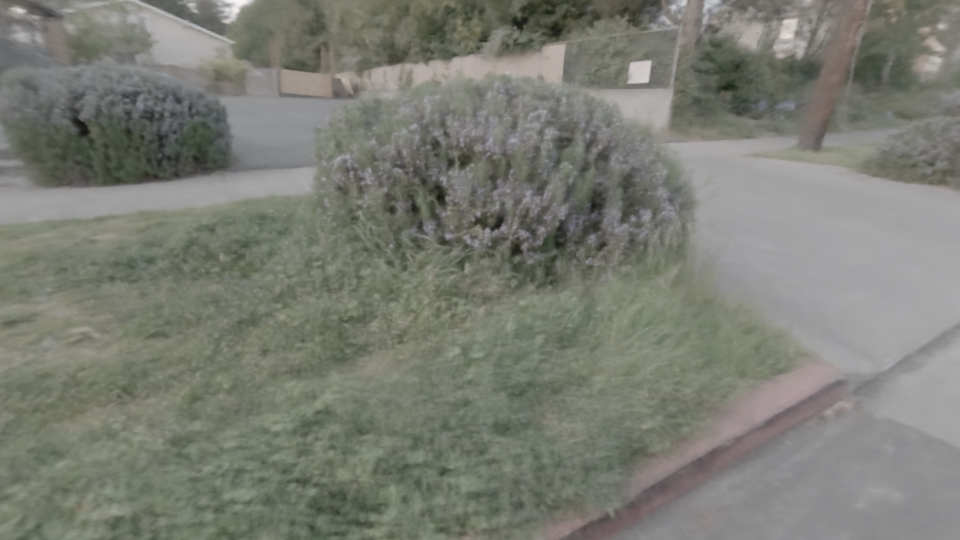} &
\includegraphics[width=0.23\linewidth]{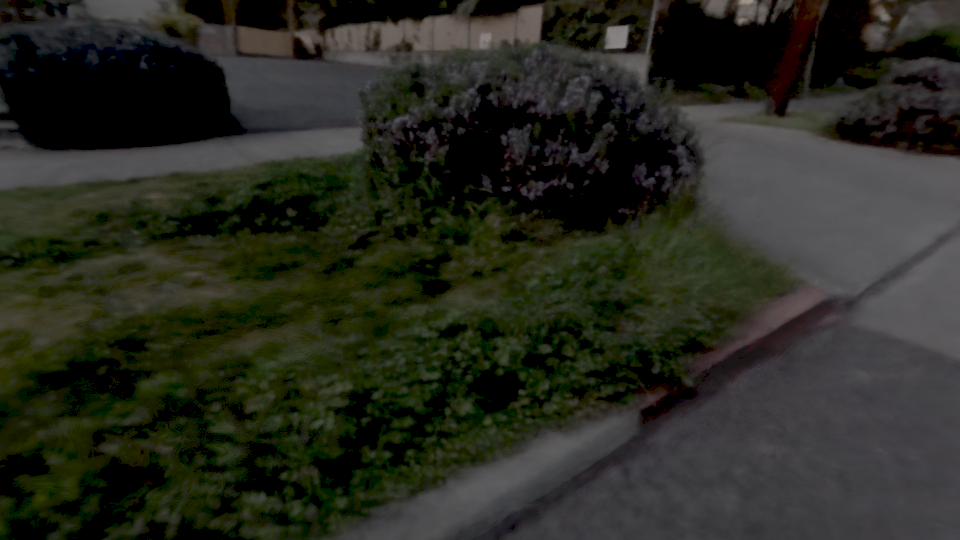} \\ \hline
Ground Truth &
\includegraphics[width=0.23\linewidth]{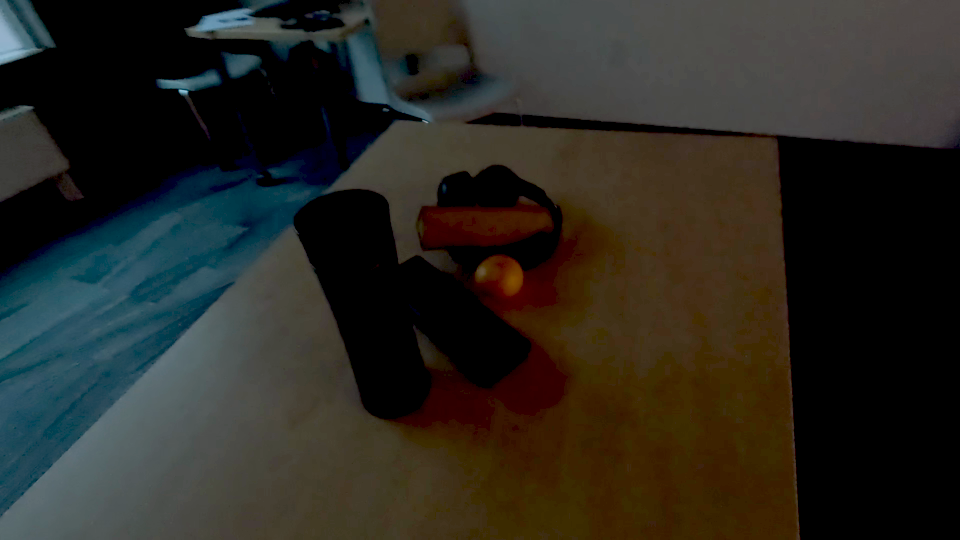} &
\includegraphics[width=0.23\linewidth]{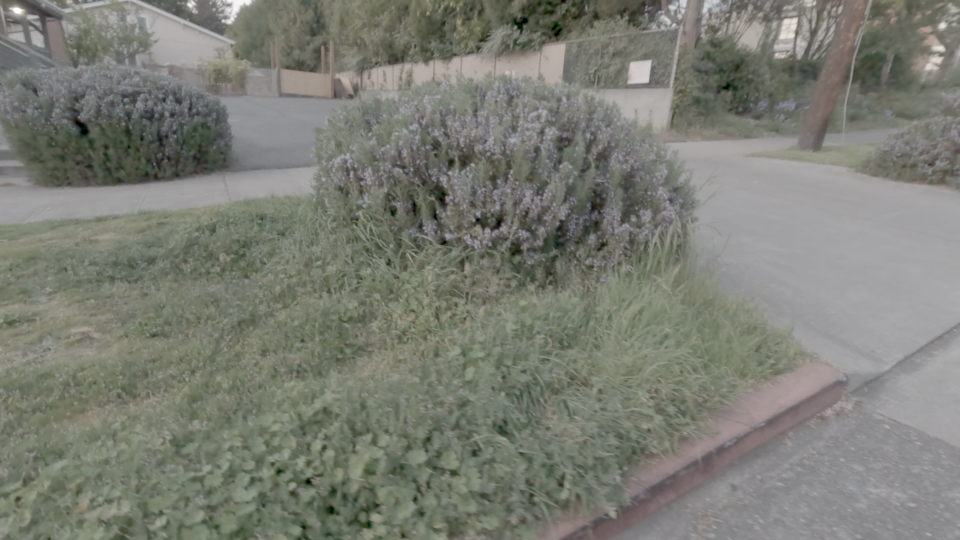} &
\includegraphics[width=0.23\linewidth]{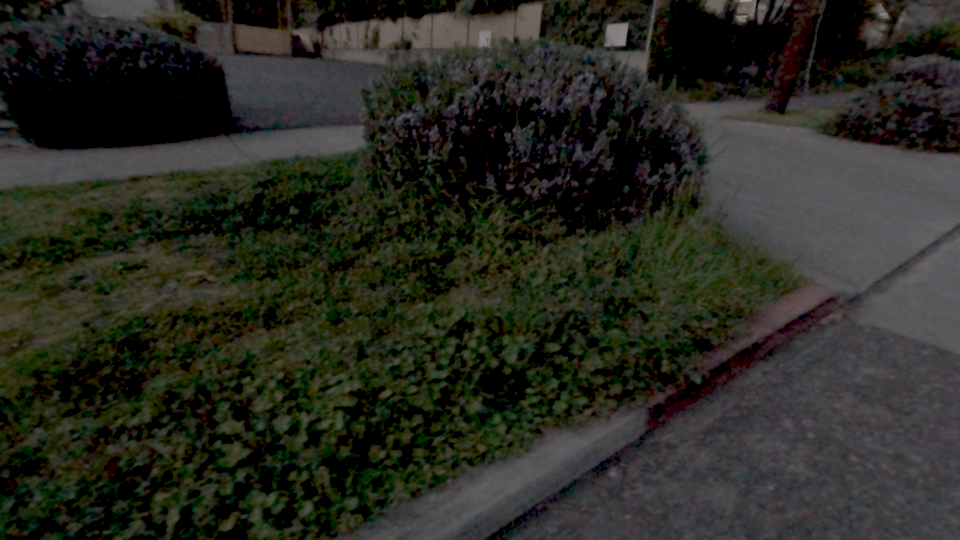} \\ \hline
\end{tabular}
\caption{RGB renderings and ground truth of three videos}
\label{tab:RGB_results}
\end{figure}

\begin{figure}[t]
\centering
\fontsize{8pt}{9pt}\selectfont
\begin{tabular}{|c|c|c|c|c|c|}
\hline
\multicolumn{1}{|c|}{\textbf{Color Space}} &
\multicolumn{1}{c|}{\textbf{GX010032}} &
\multicolumn{1}{c|}{\textbf{GX010103}} &
\multicolumn{1}{c|}{\textbf{GX010109}}\\ \hline
GPLog &
\includegraphics[width=0.23\linewidth]{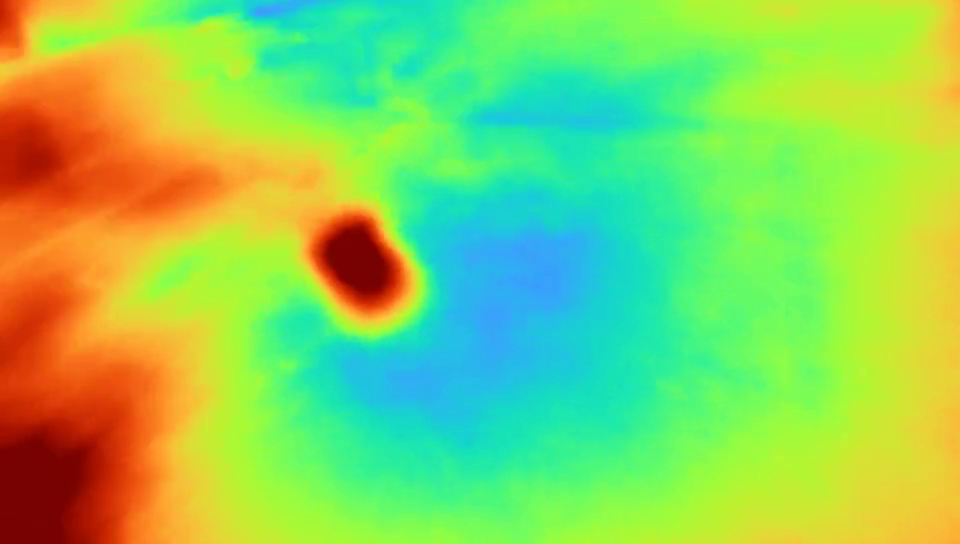} &
\includegraphics[width=0.23\linewidth]{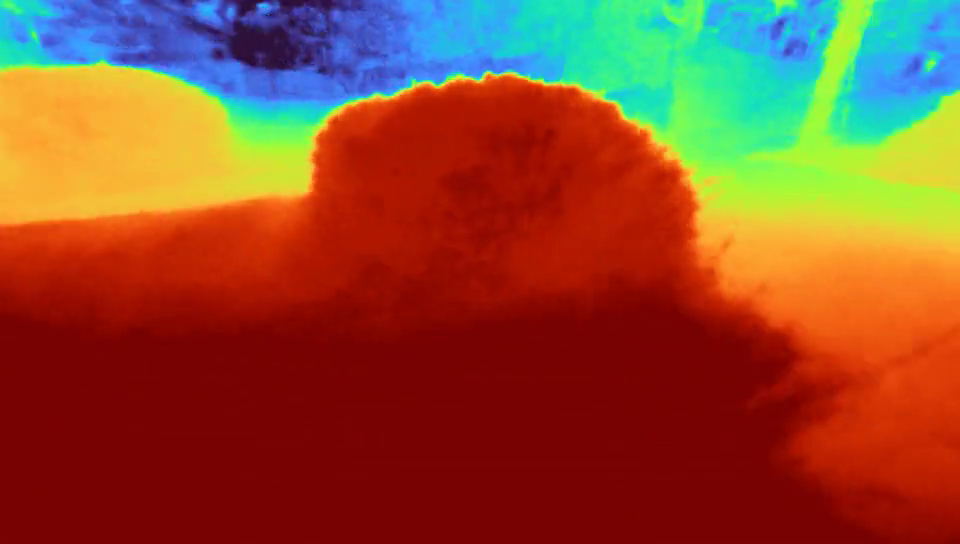} &
\includegraphics[width=0.23\linewidth]{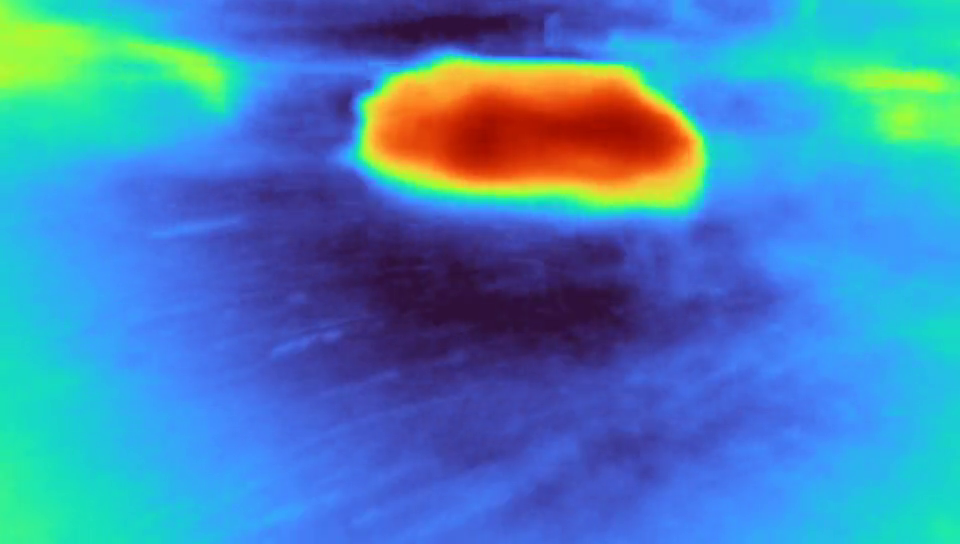} \\ \hline
linear &
\includegraphics[width=0.23\linewidth]{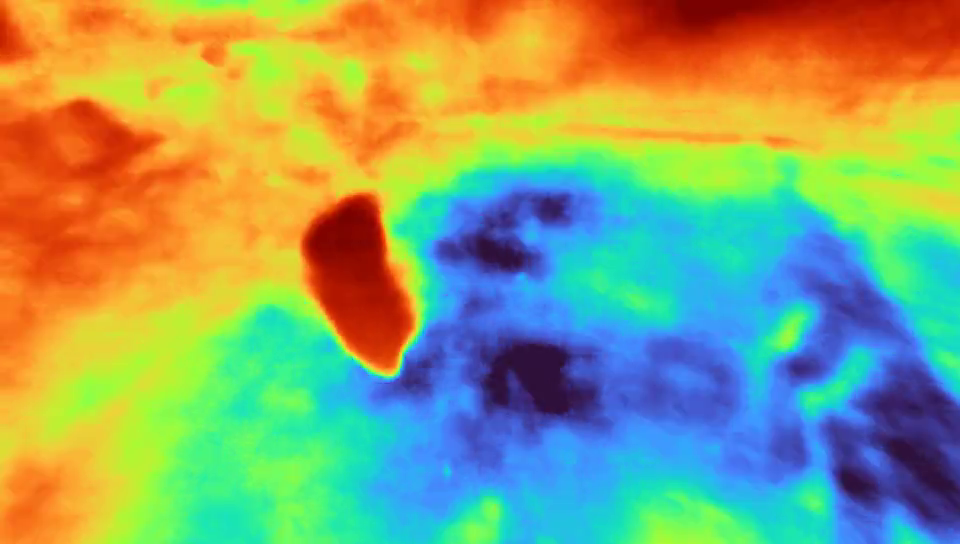} &
\includegraphics[width=0.23\linewidth]{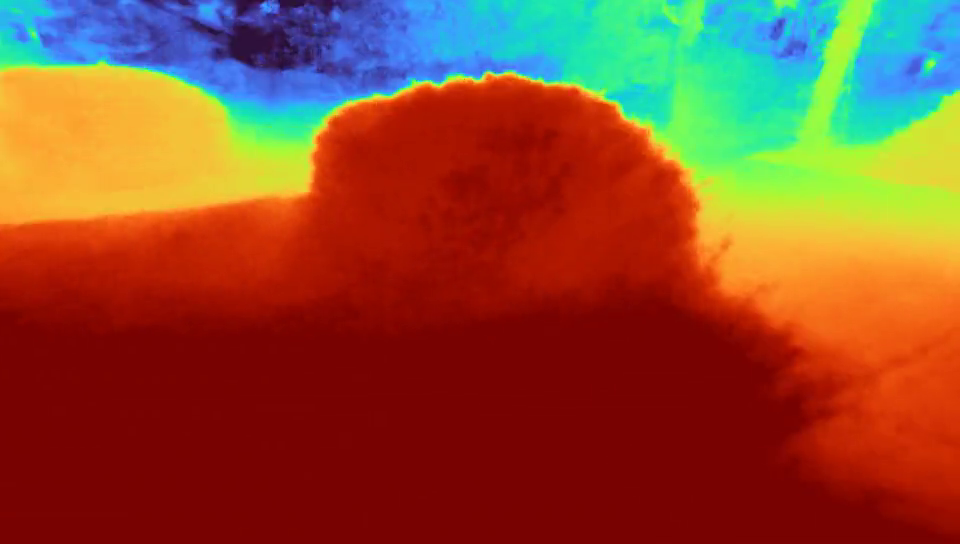} &
\includegraphics[width=0.23\linewidth]{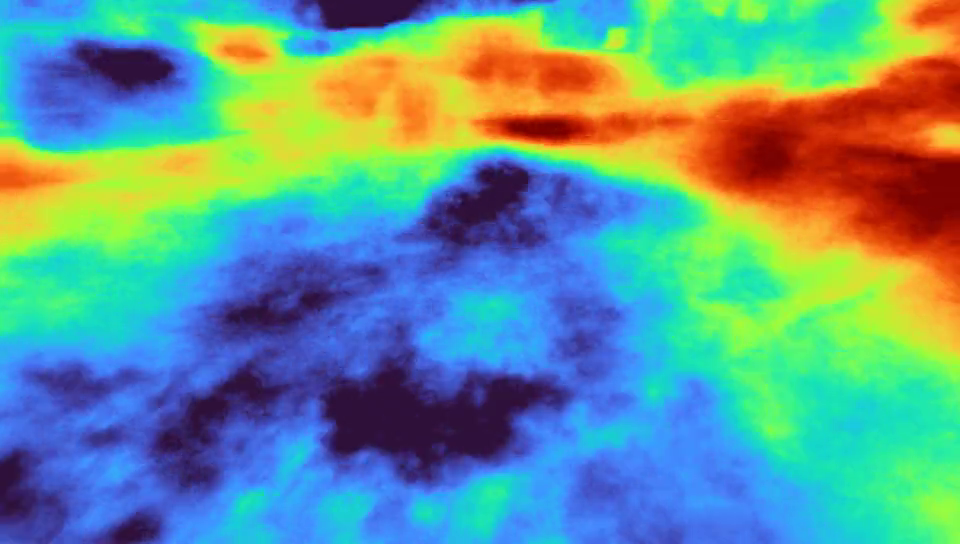} \\ \hline
sRGB &
\includegraphics[width=0.23\linewidth]{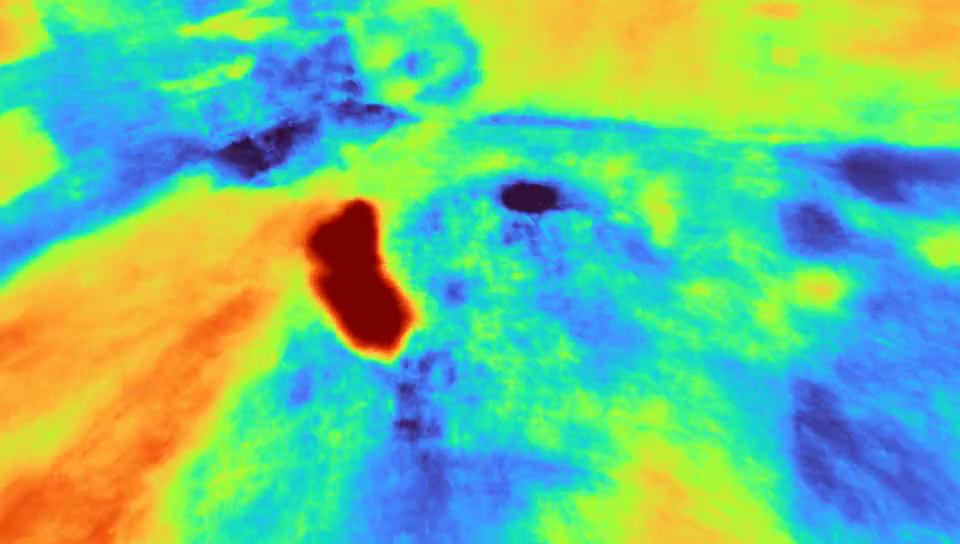} &
\includegraphics[width=0.23\linewidth]{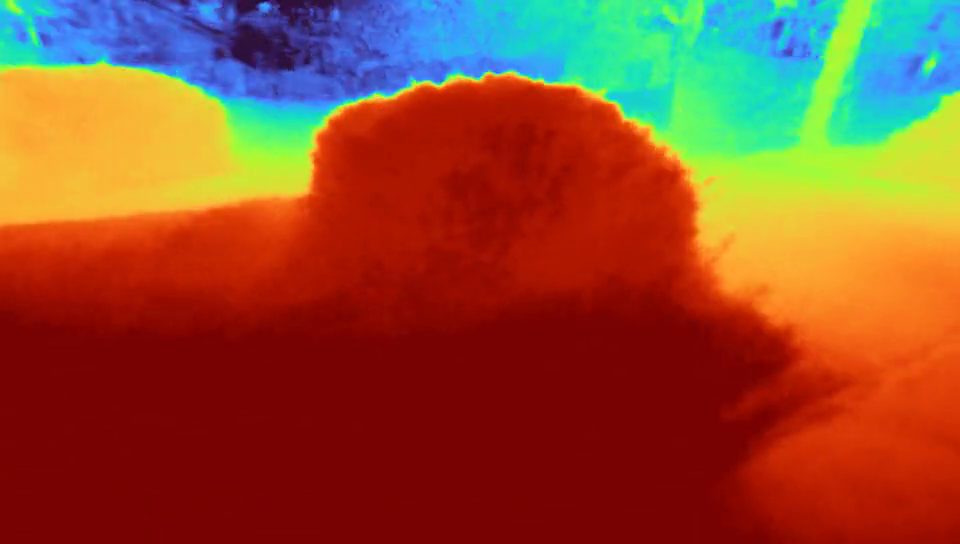} &
\includegraphics[width=0.23\linewidth]{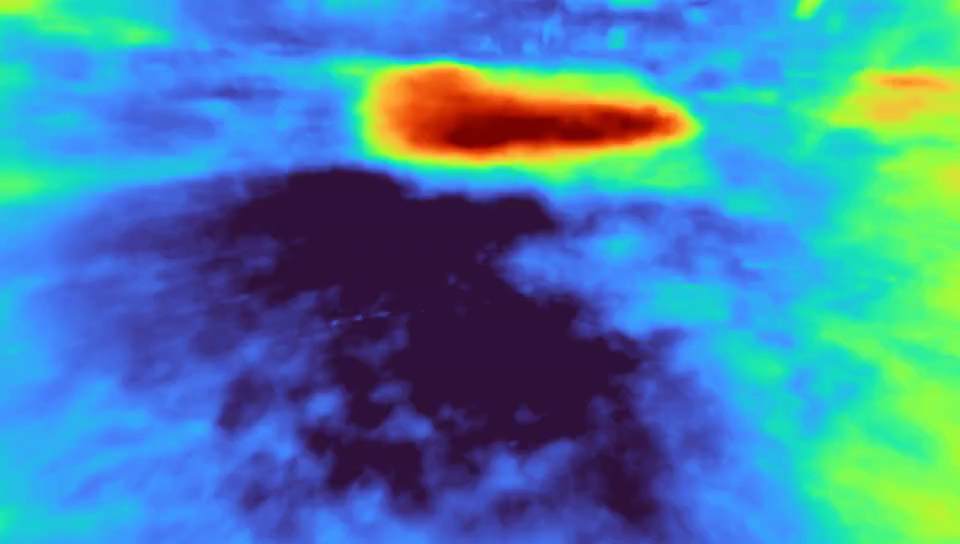} \\ \hline
TrueLog &
\includegraphics[width=0.23\linewidth]{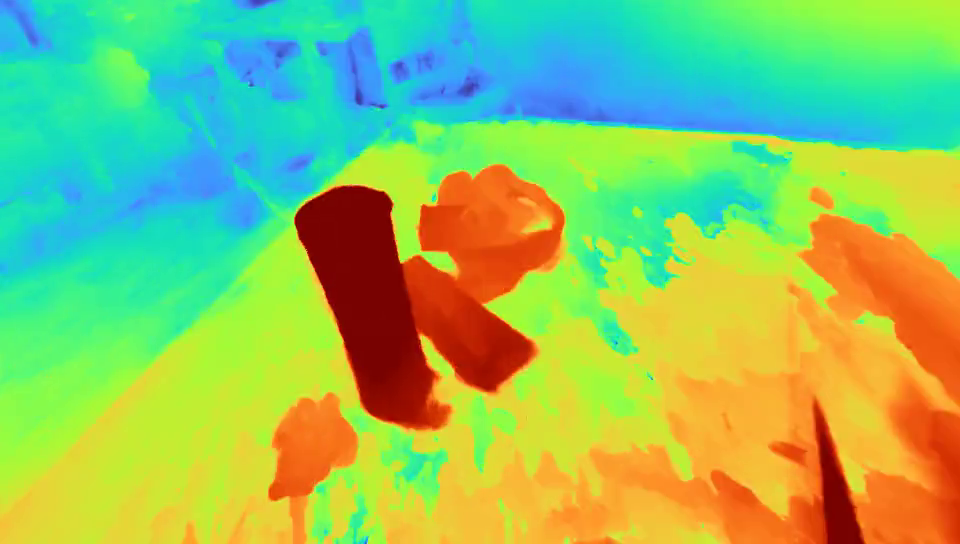} &
\includegraphics[width=0.23\linewidth]{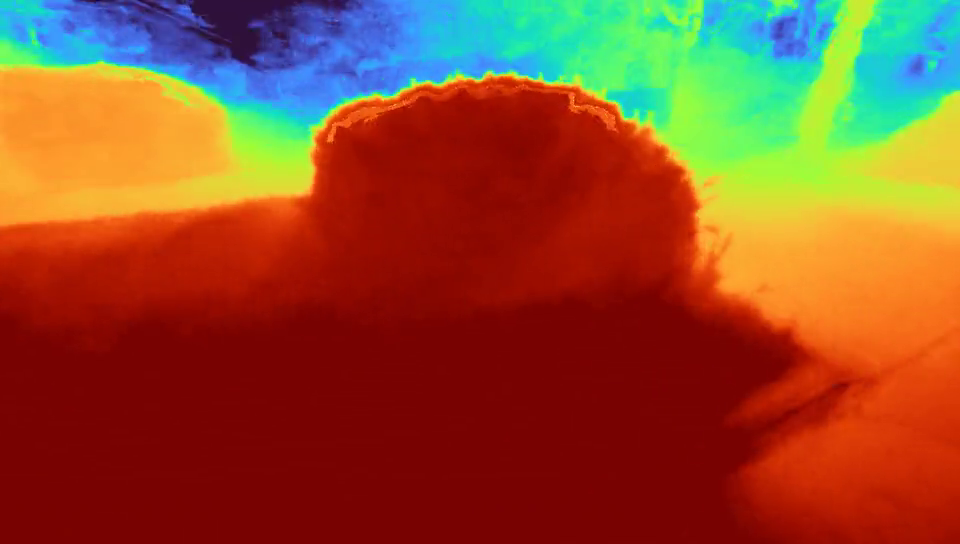} &
\includegraphics[width=0.23\linewidth]{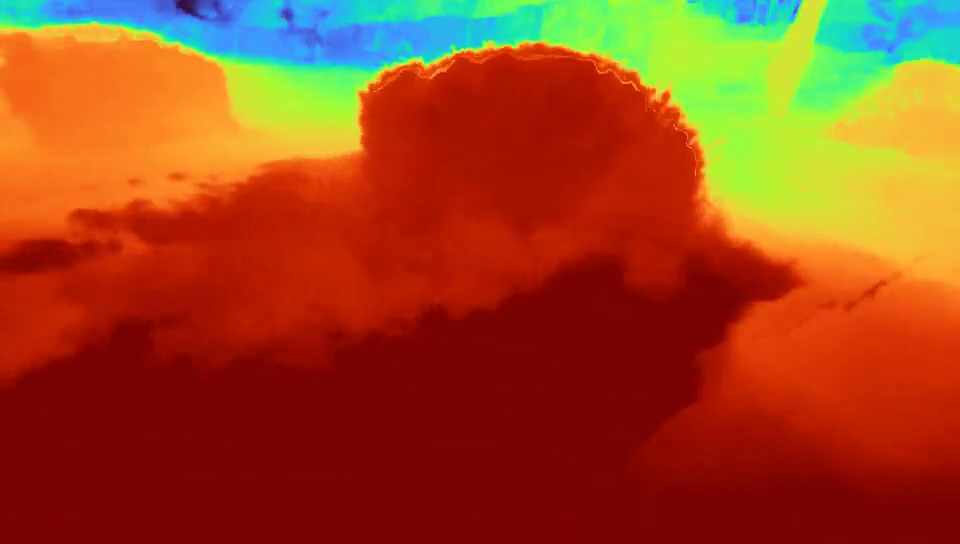} \\ \hline
Ground Truth &
\includegraphics[width=0.23\linewidth]{images/GX010032_GT.png} &
\includegraphics[width=0.23\linewidth]{images/GX010103_GT.png} &
\includegraphics[width=0.23\linewidth]{images/GX010109_GT.png} \\ \hline
\end{tabular}
\caption{Depth map renderings of three videos}
\label{tab:depth_results}
\end{figure}

The results in Table \ref{tab:Overall PSNR} are uniformly sampled from a total of 28 processed videos and ordered by the difference $PSNR_{TrueLog} - PSNR_{sRGB}$ from highest to lowest. They demonstrate consistent improvements across diverse scenes. The most significant gain is observed in one of the darkest recordings. As shown in the color renderings in Figure \ref{tab:RGB_results} and depth map visualizations Figure \ref{tab:depth_results}, the TrueLog representation preserves scene details more effectively than other color spaces under low-light conditions. Note, in particular, the reconstruction of the objects on the table in video GX010032.

Even in cases with minimal PSNR improvement, such as video GX010103 in Figure \ref{tab:RGB_results}, the TrueLog renderings reveal more visual detail, particularly in the foliage, compared to the blurrier outputs from other color spaces. The corresponding depth maps in Figure \ref{tab:depth_results} column GX010103 appear largely similar across all color spaces, indicating that the perceptual advantage of TrueLog in this case is primarily in appearance rather than geometry.

\begin{table}[t]
\centering
\fontsize{8pt}{9pt}\selectfont
\begin{tabular}{|c|c|c|c|c|c|}
\hline
Video ID & $PSNR_{GPLog}$ & $PSNR_{linear}$ & $PSNR_{sRGB}$ & $PSNR_{TrueLog}$ & $\Delta$ PSNR \\
\hline
GX010032 & 27.47       & 25.4         & 25.35      & 34.43         & 9.08                    \\
% GX010455 & 27.21       & 30.5         & 30.54      & 37.14         & 6.6                      \\
% GX010416 & 28.76       & 28.77        & 28.79      & 34.5          & 5.71                     \\
GX010055 & 34.55       & 34.56        & 34.57      & 39.54         & 4.97                     \\
% GX010100 & 30.9        & 30.88        & 30.88      & 35.65         & 4.77                     \\ 
% GX010109 & 26.67       & 23.47        & 24.61      & 29.05         & 4.44                     \\
GX010059 & 33.85       & 33.89        & 33.87      & 37.19         & 3.32                     \\
% GX010405 & 27.92       & 27.95        & 27.92      & 31.07         & 3.15                     \\
% GX010099 & 25.54       & 24.99        & 24.82      & 27.83         & 3.01                     \\
GX010108 & 26.11       & 24.11        & 24.83      & 27.55         & 2.72                     \\ 
% GX010672 & 35.77       & 35.77        & 35.69      & 38.26         & 2.57                     \\
% GX010106 & 26.5        & 23.84        & 25.73      & 28.22         & 2.49                     \\
GX010650 & 26.43       & 26.42        & 26.43      & 28.31         & 1.88                     \\ \hline
% GX010007 & 43.76       & 43.81        & 43.66      & 45.07         & 1.41                     \\
% GX010094 & 28.95       & 28.93        & 28.92      & 30.17         & 1.25                     \\ 
GX010101 & 29.72       & 29.73        & 29.74      & 30.9          & 1.16                     \\
% GX010102 & 29.86       & 29.88        & 29.89      & 31            & 1.11                     \\
% GX010015 & 42.56       & 42.77        & 42.62      & 43.69         & 1.07                     \\
GX010025 & 41.54       & 41.54        & 41.54      & 41.54         & 1.04                     \\
% GX010098 & 26.54       & 27.5         & 27.29      & 28.33         & 1.04                     \\ \hline
% GX010092 & 29.84       & 29.88        & 29.87      & 30.9          & 1.03                     \\
GX010090 & 31.08       & 31.06        & 31.08      & 32.09         & 1.01                     \\
% GX010104 & 27.41       & 28.72        & 28.71      & 29.7          & 0.99                     \\
% GX010095 & 27.71       & 27.69        & 27.68      & 28.53         & 0.85                     \\
GX010097 & 27.23       & 27.43        & 27.44      & 28.29         & 0.85                     \\ 
% GX010096 & 25.93       & 27.6         & 27.61      & 28.42         & 0.81                     \\
% GX010105 & 27.71       & 28.45        & 28.38      & 29.09         & 0.71                     \\
GX010103 & 28.19       & 28.18        & 28.2       & 28.84         & 0.64 \\                   
\hline
\end{tabular}
\caption{Overall PSNR results of captured videos (sampled)}
\label{tab:Overall PSNR}
\end{table}

%\subsection{Illumination Sensitivity and Dynamic Range}
\textbf{Illumination Sensitivity and Dynamic Range:} To study the effect of illumination, we recorded videos of the same scene at different times of day to simulate varying lighting conditions (video IDs: GX010090 to GX010109). Each video was processed using all four color spaces. If an exception occurred during training, the experiment was rerun to ensure valid results.  To quantify brightness, we computed average luminance from a randomly sampled 25\% of the frames, using the Y weighting: $I = 0.299 \cdot R + 0.587 \cdot G + 0.114 \cdot B$

\begin{figure}[tbp]
\centering
\includegraphics[height=1.5in]{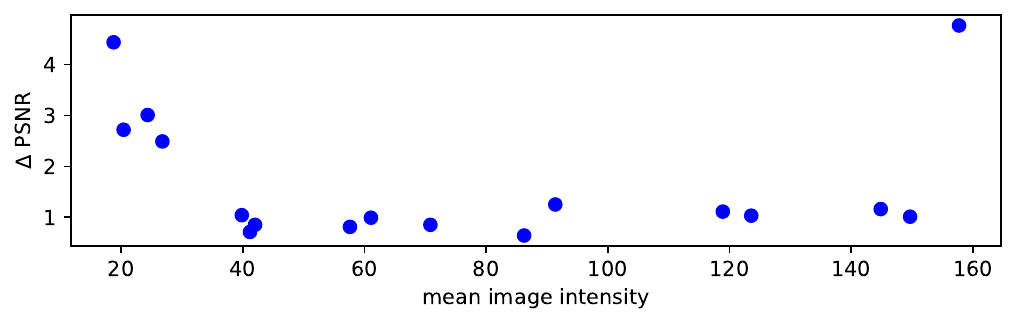}
\caption{PSNR improvement of TrueLog over sRGB as a function of scene brightness}
\label{fig:brightness}
\end{figure}

Figure \ref{fig:brightness} plots the PSNR improvement of TrueLog over sRGB as a function of scene brightness. We observe that TrueLog provides significant gains in extremely low or high illumination conditions, while improvements are more modest under well-balanced lighting. Notably, in one of the darkest scenes (GX010109), the TrueLog representation reveals much finer details in both the RGB rendering in Figure \ref{tab:RGB_results} and the depth map in Figure \ref{tab:depth_results} compared to other color spaces, similar to the types of results reported by RAWNeRF.

%\subsection{Robustness}
\textbf{Robustness:} We conducted experiments on nine videos using four different color spaces. For each video–color space pair, we ran the training process 10 times and report the mean, minimum, maximum, and standard deviation of the PSNR values. Representative results of four of the videos are summarized in Table~\ref{tab:Robustness}, with the remainder in the supplemental.

While some color spaces exhibit instability, the TrueLog representation consistently demonstrates equal or superior robustness across diverse videos and scenes. Notably, in darker videos such as GX010098, GX010099, and GX010109, TrueLog yields lower standard deviations, indicating more stable performance compared to the other color spaces.

% \begin{table}[htbp]
% \centering
% \fontsize{8pt}{9pt}\selectfont
% \begin{tabular}{l|rrrr|rrrr|rrrr|rrrr|rrrr}
% \hline
% Stat & & 10090 & & & & 10098 & & & & 10099 & & & & 10103 & & & & 10109 & & \\
% & GPL & Lin & sRGB & Log & GPL & Lin & sRGB & Log & GPL & Lin & sRGB & Log & GPL & Lin & sRGB & Log & GPL & Lin & sRGB & Log \\
% Stdev & 0.01 & 0.01 & 0.01 & 0.02 & 8.79 & 0.03 & 0.11 & 0.04 & 0.07 & 2.55 & 7.62 & 0.04 & 0.01 & 0.01 & 0.01 & 0.01 & 0.26 & 2.01 & 2.83 & 0.03 \\
% Avg & 31.08 & 31.06 & 31.07 & 32.09 & 23.43 & 27.52 & 27.42 & 28.31 & 25.56 & 22.95 & 21.46 & 27.87 & 28.20 & 28.19 & 28.20 & 28.84 & 26.45 & 23.42 & 21.83 & 29.07 \\
% Min & 31.08 & 31.05 & 31.05 & 32.06 & 0 & 27.48 & 27.19 & 28.25 & 25.45 & 17.63 & 0 & 27.83 & 28.19 & 28.18 & 28.19 & 28.83 & 25.85 & 20.11 & 16.55 & 29.03 \\
% Max & 31.09 & 31.07 & 31.09 & 32.12 & 26.78 & 27.56 & 27.50 & 28.38 & 25.67 & 25.42 & 24.75 & 27.94 & 28.21 & 28.20 & 28.21 & 28.85 & 26.68 & 25.41 & 25.01 & 29.13 \\ \hline
% \end{tabular}
% \caption{Consistency of training results across representation spaces}
% \label{tab:Robustness}
% \end{table}

\begin{table}[t]
\centering
\fontsize{8pt}{9pt}\selectfont
\begin{tabular}{|c|c|c|c|c|c|}
\hline
Video ID & Color & StdDev & Avg & Min & Max \\
\hline
\multirow{4}{*}{\textbf{GX010090}} & GPLog   & 0.01 & 31.08 & 31.08 & 31.09 \\
                                   & linear  & 0.01 & 31.06 & 31.05 & 31.07 \\
                                   & sRGB    & 0.01 & 31.07 & 31.05 & 31.09 \\
                                   & TrueLog & 0.02 & 32.09 & 32.06 & 32.12 \\
% \hline
% \multirow{4}{*}{\textbf{GX010096}} & GPLog   & 9.35 & 16.57 & 0     & 26.32 \\
%                                    & linear  & 0.01 & 27.59 & 27.58 & 27.6  \\
%                                    & sRGB    & 0.01 & 27.60 & 27.59 & 27.62 \\
%                                    & TrueLog & 0.02 & 28.39 & 28.35 & 28.42 \\
% \hline
% \multirow{4}{*}{\textbf{GX010097}} & GPLog   & 4.09 & 25.12 & 14.42 & 27.22 \\
%                                    & linear  & 0.03 & 27.43 & 27.4  & 27.48 \\
%                                    & sRGB    & 0.02 & 27.44 & 27.4  & 27.46 \\
%                                    & TrueLog & 0.04 & 28.32 & 28.26 & 28.37 \\
\hline
\multirow{4}{*}{\textbf{GX010098}} & GPLog   & 4.15 & 24.98 & 13.95 & 26.78 \\
                                   & linear  & 0.03 & 27.52 & 27.48 & 27.56 \\
                                   & sRGB    & 0.11 & 27.42 & 27.19 & 27.5  \\
                                   & TrueLog & 0.04 & 28.31 & 28.25 & 28.38 \\
\hline
\multirow{4}{*}{\textbf{GX010099}} & GPLog   & 0.07 & 25.56 & 25.45 & 25.67 \\
                                   & linear  & 2.55 & 22.95 & 17.63 & 25.42 \\
                                   & sRGB    & 4.09 & 22.60 & 11.36 & 24.75 \\
                                   & TrueLog & 0.04 & 27.87 & 27.83 & 27.94 \\
% \hline
% \multirow{4}{*}{\textbf{GX010101}} & GPLog   & 0.01 & 29.72 & 29.71 & 29.73 \\
%                                    & linear  & 0.01 & 29.73 & 29.72 & 29.74 \\
%                                    & sRGB    & 0.01 & 29.72 & 29.71 & 29.73 \\
%                                    & TrueLog & 0.01 & 30.89 & 30.88 & 30.9  \\
% \hline
% \multirow{4}{*}{\textbf{GX010102}} & GPLog   & 0.01 & 29.88 & 29.86 & 29.89 \\
%                                    & linear  & 0.01 & 29.87 & 29.85 & 29.89 \\
%                                    & sRGB    & 0.01 & 29.89 & 29.88 & 29.9  \\
%                                    & TrueLog & 0.02 & 30.99 & 30.95 & 31.03 \\
% \hline
% \multirow{4}{*}{\textbf{GX010103}} & GPLog   & 0.01 & 28.20 & 28.19 & 28.21 \\
%                                    & linear  & 0.01 & 28.19 & 28.18 & 28.2  \\
%                                    & sRGB    & 0.01 & 28.20 & 28.19 & 28.21 \\
%                                    & TrueLog & 0.01 & 28.84 & 28.83 & 28.85 \\
\hline
\multirow{4}{*}{\textbf{GX010109}} & GPLog   & 0.26 & 26.45 & 25.85 & 26.68 \\
                                   & linear  & 2.01 & 23.42 & 20.11 & 25.41 \\
                                   & sRGB    & 2.83 & 21.83 & 16.55 & 25.01 \\
                                   & TrueLog & 0.03 & 29.07 & 29.03 & 29.13 \\
\hline
\end{tabular}
\caption{Consistency of training results across representation spaces}
\label{tab:Robustness}
\end{table}

%\subsection{Effect of Training Iterations}
\textbf{Effect of Training Iterations:} We analyzed how training duration affects model performance across color spaces by training two representative videos (GX010106 and GX010405) using seven iteration settings: 500, 1000, 2000, 5000, 10000, 20000, and 25000. For one video, we trained to 30000 iterations but observed negligible change ($\sim 0.01$ dB PSNR). %During our 48 total experimental runs, we encountered exceptions in two cases, which were successfully resolved through retraining.

\begin{figure}[t]
\centering
\begin{tabular}{cc}
\includegraphics[height=1.25in]{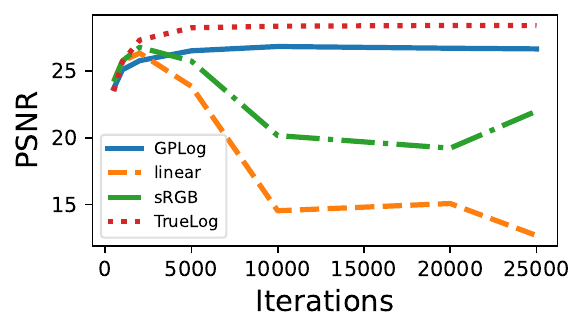} &
\includegraphics[height=1.25in]{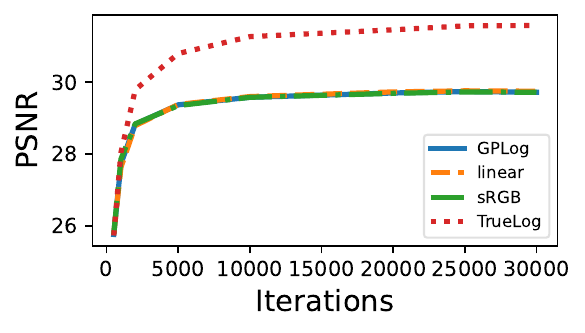} \\
(A) & (B) \\
\end{tabular}
\caption{PSNR values across training iterations for videos (A) GX010106 \& (B) GX010405. }
\label{fig:iterations_106_405}
\end{figure}

Figure \ref{fig:iterations_106_405} demonstrates distinct convergence patterns across color spaces. TrueLog outperforms others throughout the training process, with stable improvement across iterations. GPLog maintains relatively stable performance, but the linear and sRGB color spaces exhibit instability in the second dataset (GX010405), with degradation after initial peaks.

Our findings suggest that TrueLog provides the optimal balance between computational efficiency and reconstruction quality, with particularly strong performance emerging at approximately 5000 iterations. This represents an important efficiency finding, as it indicates that high-quality results can be achieved with fewer iterations than previously assumed.

%\subsection{Specialness of Log}
\textbf{Specialness of Log:} We evaluated alternative scaled logarithmic conversions across four videos to examine the uniqueness of TrueLog. Specifically, we evaluated (\ref{eq:scaledlog}) using $k=255\times100$ (Log100) and $k=255\times0.1$ (Log01).
\begin{equation}
    ScaledLog(x)=\frac{log(kx+1)}{log(k+1)}
    \label{eq:scaledlog}
\end{equation}

As shown in Table \ref{tab:scaled_log}, TrueLog consistently outperforms similar logarithmic transformations with different scaling coefficients. The performance gap demonstrates that log space is special for this task, and it does not generalize to other similar transformations.
\begin{table}[tbp]
\centering
\begin{tabular}{|c|c|c|c|}
\hline
\textbf{Video} & \textbf{TrueLog} & \textbf{Log100} & \textbf{Log01} \\
\hline
GX010032 & 38.06 & 26.41 & 26.67 \\
GX010099 & 27.83 & 25.27 & 12.23 \\
GX010109 & 29.05 & 24.93 & 27.03 \\
GX010405 & 31.07 & 27.75 & 27.93 \\
\hline
\end{tabular}
\caption{PSNR results comparing TrueLog against two scaled log conversions}
\label{tab:scaled_log}
\end{table}

\begin{figure}[t]
\centering
\begin{tabular}{cc}
\includegraphics[height=1.25in]{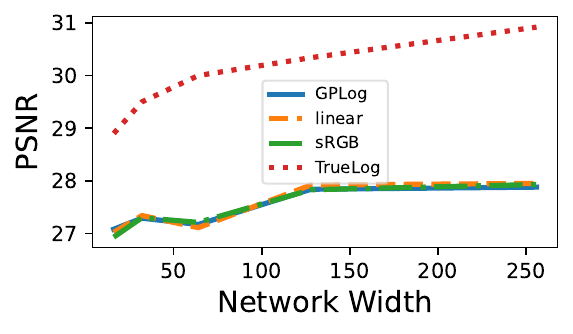} & 
\includegraphics[height=1.25in]{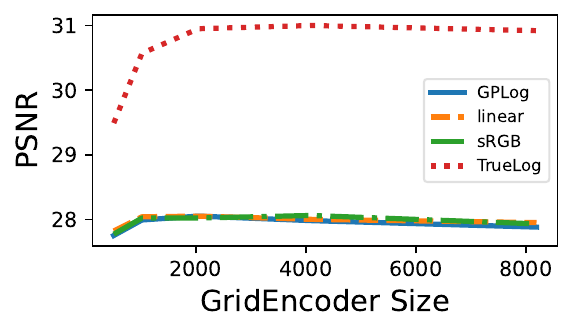} \\
(A) &  
(B) \\
\end{tabular}
\caption{Network complexity versus PSNR for video GX010405. (A) shows the impact of changes in network width, (B) shows the impact of changes in the grid encoder size.}
\label{fig:netwidth_gridencoder_GX010405}
\end{figure}

\begin{figure}[ht]
\centering
\begin{tabular}{cc}
\includegraphics[width=0.47\linewidth]{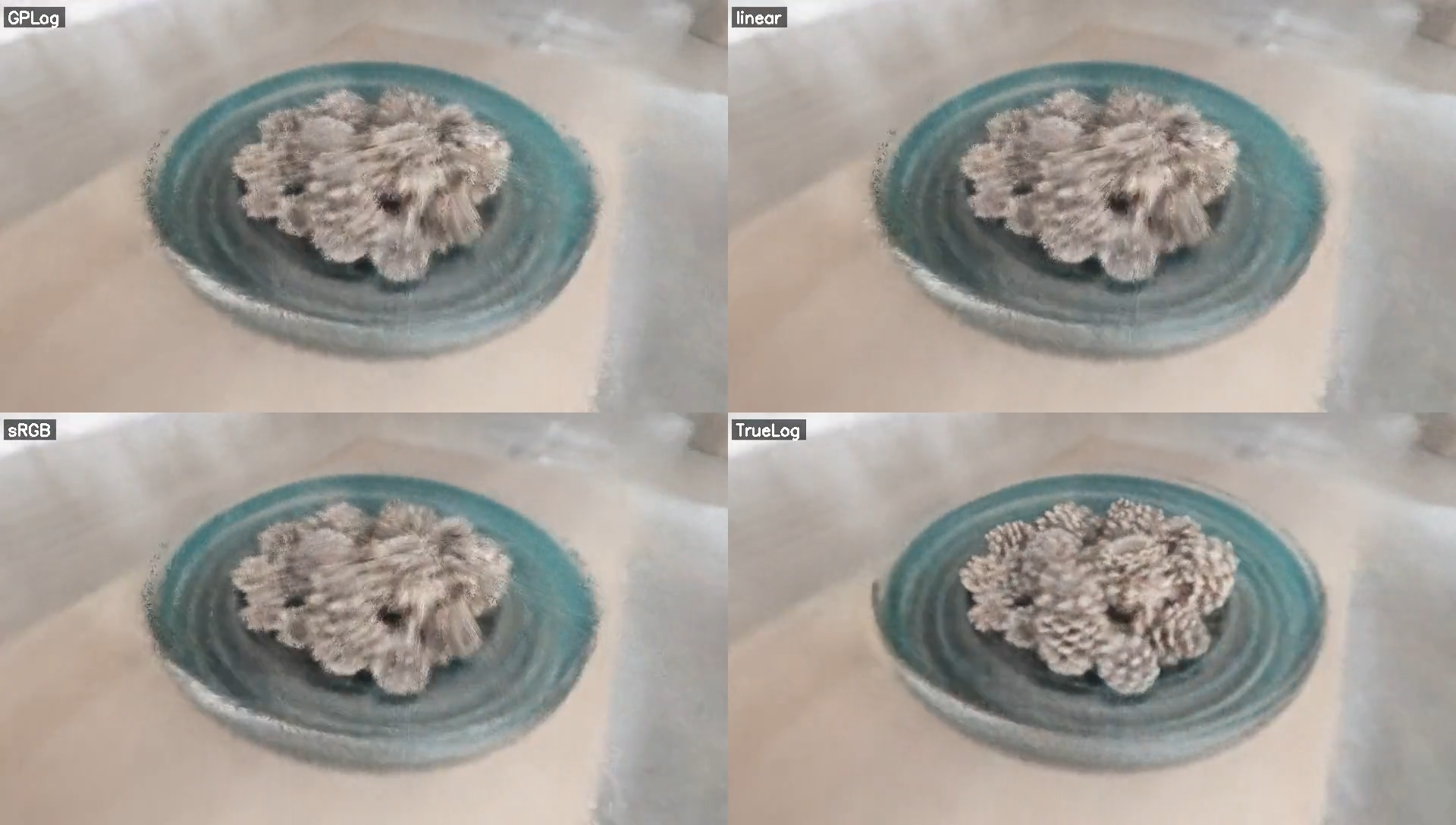} & 
\includegraphics[width=0.47\linewidth]{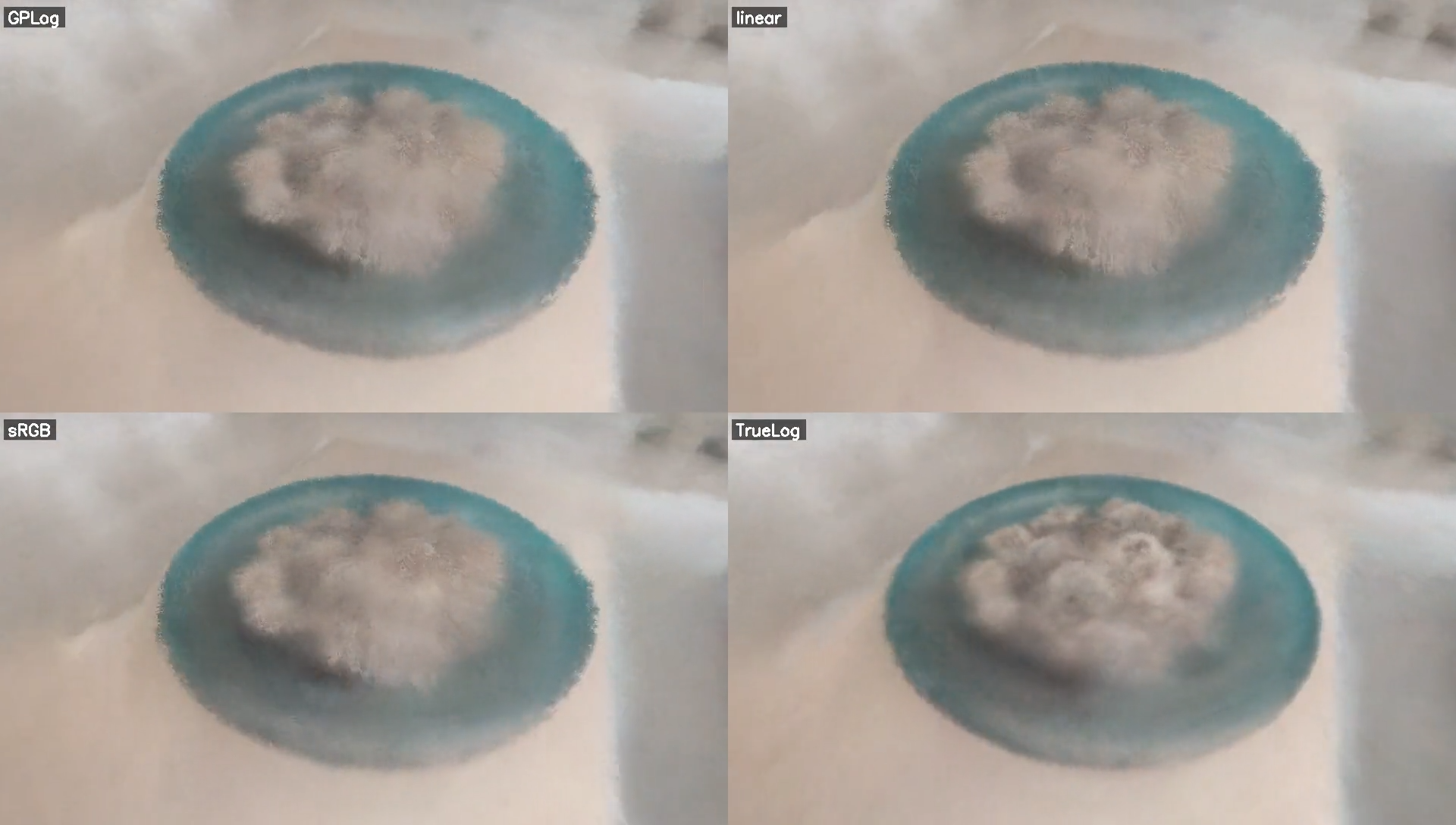} \\
Mip-NeRF & Robust NeRF \\
\end{tabular}
\caption{Video GX010405 trained with different NeRF approaches}
\label{fig:mipnerf_robustnerf_405}
\end{figure}

%\subsection{Network Compactness Analysis}
\textbf{Network Compactness Analysis:} We evaluated compactness using the BiLaRF architecture by reducing MLP widths from 256 to 16 neurons (Figure \ref{fig:netwidth_gridencoder_GX010405} A). TrueLog consistently outperformed other color spaces by 2-3 dB PSNR, demonstrating stronger scaling with increased width. Grid encoder size reduction experiments from 8192 to 256) revealed that while GPLog, linear, and sRGB maintained around 28 dB performance across all configurations, TrueLog scaled more effectively, reaching optimal performance (around 31 dB) at size 2048 (Figure \ref{fig:netwidth_gridencoder_GX010405} B) .
Despite longer training times and less refined outputs compared to BiLaRF, TrueLog consistently produced sharper, less blurry results in both Mip-NeRF \cite{barron2022mipnerf360} and Robust-NeRF \cite{Sabour_2023_CVPR} (Figures \ref{fig:mipnerf_robustnerf_405}) \cite{multinerf2022}. This result shows the potential generalization of TrueLog representation to other NeRF implementations.

\section{Discussion and Summary}

Our experiments show that using a log representation space during NeRF training yields consistent improvement over other color spaces, including linear, sRGB, and GPLog. Quantitatively, TrueLog improves PSNR by an average of 10\% across various scenes for the same data and network. Qualitatively, it produces sharper edges, clearer visualizations, and more accurate depth maps, particularly in challenging lighting conditions. While the BiLaRF framework provides a strong foundation for efficient training and fast convergence, incorporating log-space representations further enhances the network’s performance and stability.

For future work, we plan to extend our investigation of log-space generalization to a broader range of NeRF architectures and tasks. Although preliminary tests on mip-NeRF and robust-NeRF suggest that TrueLog maintains its advantages, exploring other approaches could provide additional insights. Additionally, being able to directly visualize the network’s output in its native color space before sRGB conversion would be valuable for analyzing how different representations influence learning dynamics.

\bibliography{bibliography-master}
\end{document}